\documentclass[letterpaper]{article} 
\usepackage{aaai25}  
\usepackage{times}  
\usepackage{helvet}  
\usepackage{courier}  
\usepackage[hyphens]{url}  
\usepackage{graphicx} 
\usepackage{natbib}  
\usepackage{caption} 
\usepackage{algorithm}
\usepackage{algorithmic}
\usepackage{amsmath}
\usepackage{amssymb}
\usepackage{booktabs}
\usepackage{multirow}
\usepackage{newfloat}
\usepackage{listings}
\DeclareCaptionStyle{ruled}{labelfont=normalfont,labelsep=colon,strut=off} 
\floatstyle{ruled}
\newfloat{listing}{tb}{lst}{}
\floatname{listing}{Listing}
\title{MergeNet: Knowledge Migration across Heterogeneous Models, Tasks, and Modalities}
\author{
    Kunxi Li\textsuperscript{\rm 1}\thanks{These authors contributed equally.},
    Tianyu Zhan\textsuperscript{\rm 1}\footnotemark[1],
    Kairui Fu\textsuperscript{\rm 1}\footnotemark[1],
    Shengyu Zhang\textsuperscript{\rm 1}\thanks{Corresponding author.},
    Kun Kuang\textsuperscript{\rm 1},\\
    Jiwei Li\textsuperscript{\rm 1},
    Zhou Zhao\textsuperscript{\rm 1},
    Fan Wu\textsuperscript{\rm 2},
    Fei Wu\textsuperscript{\rm 1}
}
\affiliations{
    \textsuperscript{\rm 1}Zhejiang University \\
    \textsuperscript{\rm 2}Shanghai Jiao Tong University \\
    \{kunxili,3200105641,fukairui.fkr, sy\_zhang, kunkuang, jiwei\_li, zhaozhou, wufei\}@zju.edu.cn,fwu@cs.sjtu.edu.cn

}

\usepackage{bibentry}

\begin{document}

\maketitle

\begin{abstract}
In this study, we focus on heterogeneous knowledge transfer across entirely different model architectures, tasks, and modalities. Existing knowledge transfer methods (e.g., backbone sharing, knowledge distillation) often hinge on shared elements within model structures or task-specific features/labels, limiting transfers to complex model types or tasks. To overcome these challenges, we present MergeNet, which learns to bridge the gap of parameter spaces of heterogeneous models, facilitating the direct interaction, extraction, and application of knowledge within these parameter spaces. The core mechanism of MergeNet lies in the parameter adapter, which operates by querying the source model's low-rank parameters and adeptly learning to identify and map parameters into the target model. MergeNet is learned alongside both models, allowing our framework to dynamically transfer and adapt knowledge relevant to the current stage, including the training trajectory knowledge of the source model. Extensive experiments on heterogeneous knowledge transfer demonstrate significant improvements in challenging settings, where representative approaches may falter or prove less applicable.
\end{abstract}

%

\section{Introduction}

\begin{figure}[t]
  \centering
  \includegraphics[width=7.61cm]{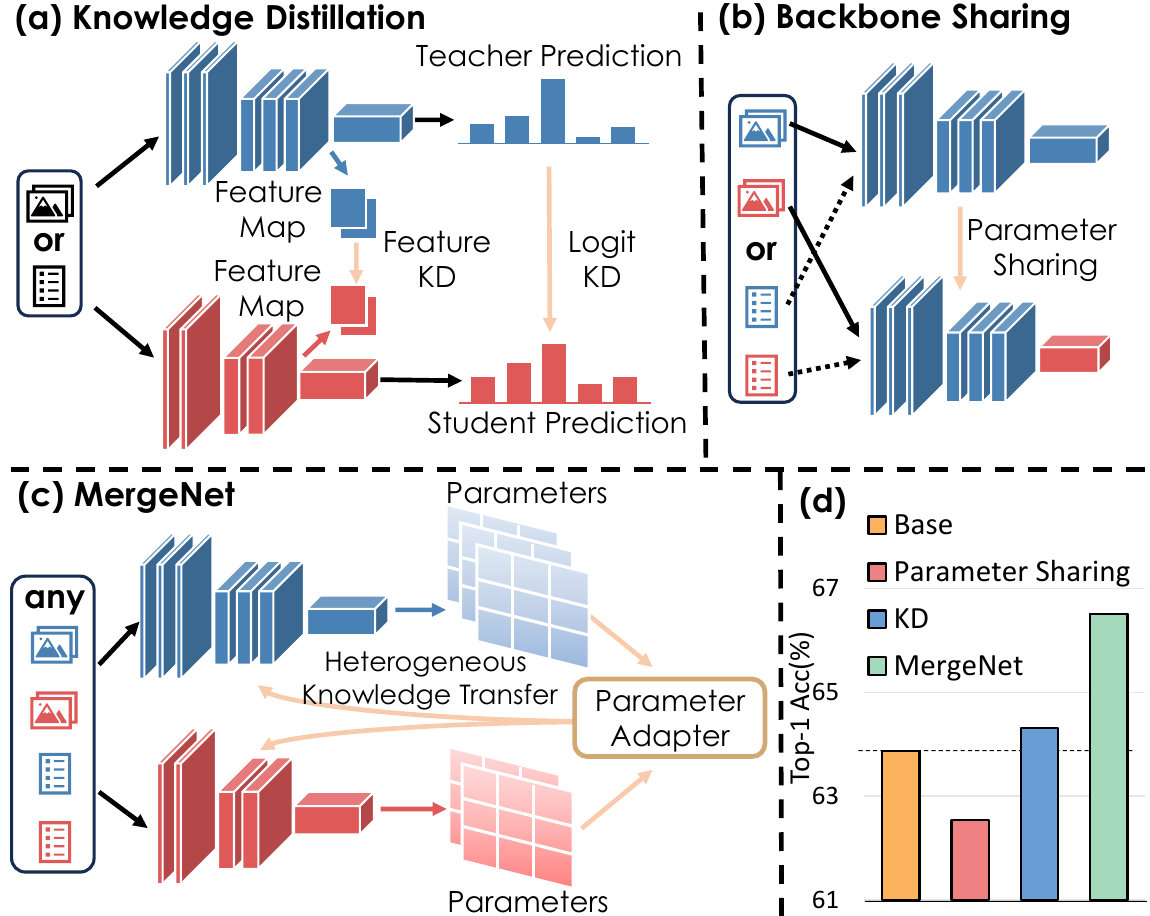}
  \caption{(a)-(c): Compare knowledge distillation, backbone sharing, and our proposed MergeNet. The orange arrows represent the flow of knowledge. (d): The parameter sharing method is ineffective for heterogeneous knowledge transfer, and in fact, may lead to a loss of accuracy due to the incompatibility of knowledge.}
  \label{fig:histogram}
\end{figure}

In an era where edge computing devices are ubiquitous, the deployment of deep neural networks (DNNs) on such devices is often constrained by limited computational resources and storage capacity. This limitation typically necessitates the utilization of smaller neural network architectures, which, while computationally economical, often compromise performance. A promising approach to mitigate this limitation involves the strategic transfer of knowledge from larger, more capable models to these constrained counterparts. A prime  example of this approach is knowledge distillation (KD) \cite{hinton2015distilling,beyer2022knowledge,shen2021label}. This technique involves training a compact student model to mimic the output logic or intermediate layer representations of a more comprehensive teacher model, thereby improving its accuracy. 
Transfer learning \cite{zhuang2020comprehensive,he2021towards,hassanpour2023survey} offers another approach, commonly through the pre-training and fine-tuning workflow, where knowledge learned during pre-training on large-scale datasets is applied to downstream tasks via backbone sharing. However, these methods typically require similar task types or model architectures, which can limit their applicability. For instance, in diverse Internet of Things (IoT) applications, devices often face varying computational resources and tasks, making it difficult to find matching models for knowledge transfer. Therefore, facilitating \textbf{\emph{heterogeneous knowledge transfer}} across different model architectures, tasks, and modalities remains a pressing challenge that needs to be addressed.

Unlike previous methods, we consider knowledge transfer between models from a different perspective. We focus on the inherent properties of model parameters,  regarding them as the natural carriers of knowledge. 
An intuitive method is to adopt the idea of parameter sharing, similar to \citet{cai2021network} by putting the smaller model into the larger one, making it a subset of the larger model for additional supervision. We test the effectiveness of parameter sharing on two models with different architectures, ResNet50 \cite{he2016deep} and MobileNetV2 \cite{sandler2018mobilenetv2}, on the CIFAR-100 \cite{krizhevsky2009learning}. As shown in Figure \ref{fig:histogram}(d), the performance of MobileNetV2 suggests that direct parameter sharing may not be the optimal conduit for heterogeneous knowledge transfer. We speculate that the reasons might be: (i) Direct parameter sharing might disrupt the knowledge of the original modules when heterogeneous modules have significantly different functionalities, like sharing between linear layers and attention mechanism modules. (ii) Typically, larger models contain more advanced knowledge (\textit{e.g.}, fine-grained feature combination) than smaller ones, which the latter might not directly comprehend, leading to potential incompatibility in knowledge between models due to direct parameter sharing.

In this paper, we propose Knowledge \underline{\textbf{M}}igration across H\underline{\textbf{e}}te\underline{\textbf{r}}o\underline{\textbf{g}}eneous Models, Tasks, and Modaliti\underline{\textbf{e}}s, abbreviated as \textbf{MergeNet}, which is a universal framework for knowledge transfer. To address the issue of knowledge incompatibility between heterogeneous models, we introduce parameter adapter between the source and target models to refine and summarize the knowledge within the parameters. Specifically, we re-encode model parameters through low-rank decomposition, obtaining manageable low-rank matrices that package the comprehensive knowledge of the original parameters. 
Next, using Low-rank Parametric Knowledge Adapter (LPKA), we facilitate interactions within the low-rank parameter space and generate parameters that contain knowledge from both models, thus achieving knowledge transfer. This process can be likened to the target model extracting the knowledge it currently needs from the source model, based on its existing knowledge.

To facilitate a comprehensive analysis of the heterogeneous knowledge transfer capabilities of our method, we conduct extensive experiments across various challenging scenarios, including knowledge transfer between entirely different model architectures (\textit{e.g.}, ResNet50 $\leftrightarrow$ MobileNetV2), tasks (\textit{e.g.}, Classification $\leftrightarrow$ Question Answering), and modalities (\textit{e.g.}, Image $\leftrightarrow$ Text). The results demonstrate that our method significantly outperforms traditional knowledge transfer approaches, effectively transferring knowledge across diverse scenarios and showcasing exceptional potential.

Overall, our contributions can be summarized as follows:

\begin{itemize}
  \item 
  We introduce a novel method which leverage parameters as the medium to achieve knowledge transfer between heterogeneous models, tasks and modalities.
  \item 
  Our proposed MergeNet not only extracts aligned and effective information from network parameters but also provides an efficient transfer with less computation.
  \item Our extensive experiments across various challenging scenarios validate the effectiveness of MergeNet. The results clearly demonstrate that MergeNet significantly improves model performance and surpasses the widely-used knowledge distillation techniques.
\end{itemize}

\section{Related Work}
In the evolving field of artificial intelligence, knowledge transfer stands out as a crucial mechanism, facilitating the application of insights from specific tasks, domains, or models to new and related contexts \cite{fu2024diet}. Transfer learning \cite{zhuang2020comprehensive,he2021towards} is one of the most prevalent forms, involving the extraction of knowledge from one task and applying it to a different but related task to boost the model's generalization and learning efficiency. This often entails pre-training a model on extensive datasets or complex tasks, followed by fine-tuning it for more specific tasks. Domain adaptation \cite{farahani2021brief,hassanpour2023survey,ding2023deep,kim2021domain} focuses on adapting models trained on one or multiple source domains to function effectively in a distinct target domain. Multi-task learning \cite{crawshaw2020multi,standley2020tasks,kurin2022defense,groenendijk2021multi} trains models to solve multiple related tasks simultaneously, promoting cross-task information flow through shared representations. This approach allows each task to benefit from features and patterns learned from other tasks. These cross-task learning methods not only improve data utilization and learning efficiency but also enhance the adaptability and robustness of models when confronted with new tasks.

Knowledge distillation (KD) \cite{hinton2015distilling,beyer2022knowledge,shen2021label} is a key knowledge transfer method that involves transferring knowledge from large teacher networks to smaller student models. This is achieved by training the student model to replicate the output logic \cite{hinton2015distilling,zhao2022decoupled,yang2021knowledge,zhou2021rethinking} or the intermediate layer features \cite{zagoruyko2016paying,yang2022focal} of the teacher model. Recent studies have focused on optimizing the distillation process by improving loss functions and strategies to boost the student's performance. For instance, MGD \cite{yang2022masked} enhances the transfer process by masking student model features, forcing it to reconstruct the teacher's characteristic features for a more accurate and robust knowledge transfer. Similarly, NKD \cite{yang2023knowledge} enhances the student model's use of soft labels by normalizing non-target logits, allowing the student to better utilize the rich information embedded in these labels.

Unlike previous knowledge transfer methods, our approach is designed to address scenarios of heterogeneous knowledge transfer across entirely different model architectures, tasks, and modalities. Technically, our method does not rely on output logic or intermediate layer features specific to any task or model architecture. Instead, it utilizes universal model parameters as carriers of knowledge. By bridging the differences in parameter spaces of heterogeneous models, our method fuses knowledge within the parameter space and maps it onto the parameters of the target model.

\begin{figure*}[h]
  \centering
  \includegraphics[width=175mm]{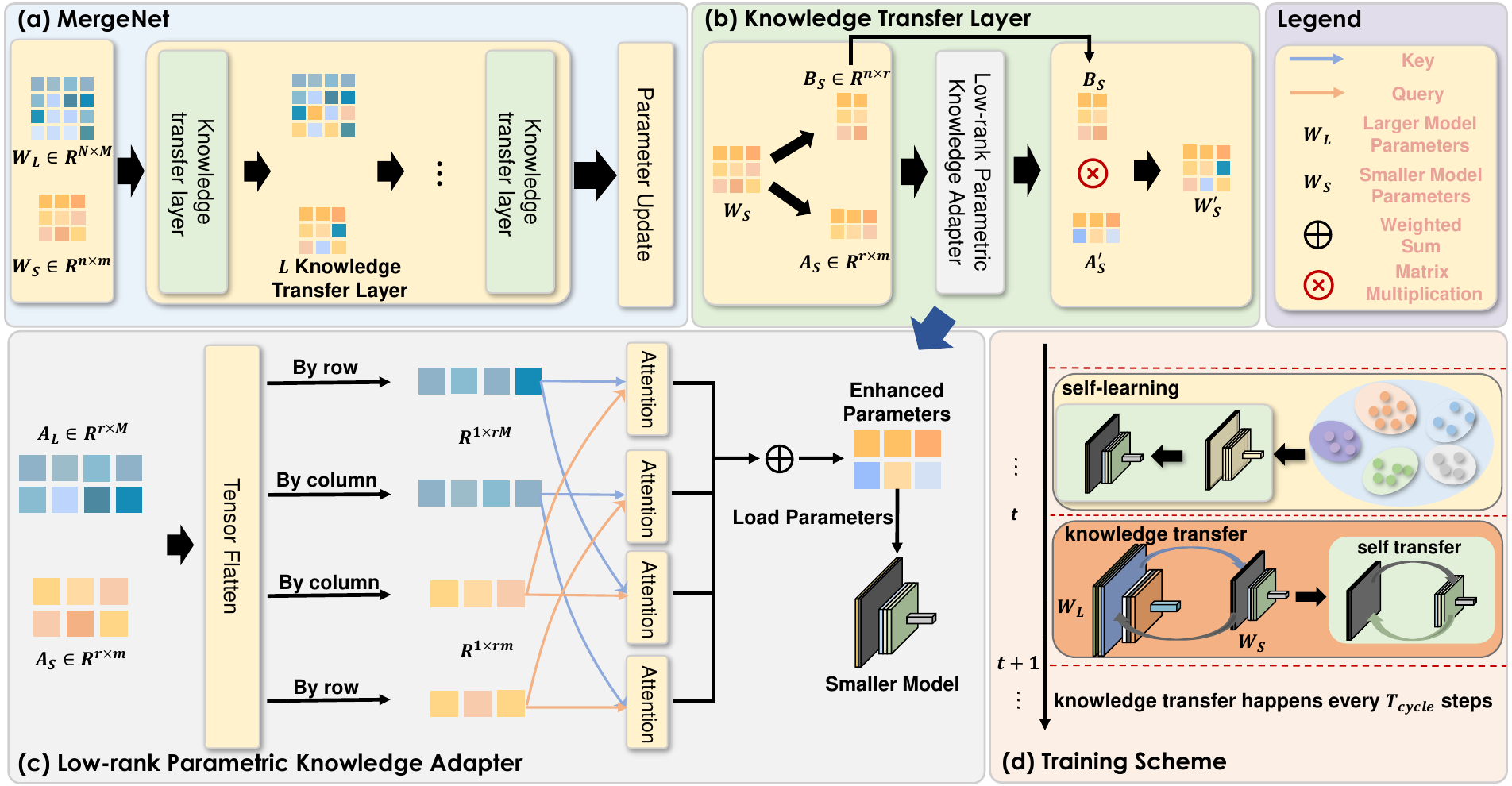}
  \caption{Overview of MergeNet. MergeNet takes parameters from different models as inputs and generates parameters that integrate knowledge from these models, where more knowledge transfer layers indicate a greater amount of knowledge transferred. It is important to note that the descriptions in (b) and (c) are based on the knowledge transfer from larger model to smaller model, but the process from smaller model to larger model is completely symmetrical.}
  \label{fig:model}
\end{figure*}

\section{Method}
\subsection{Problem Formulation}
The objective of heterogeneous knowledge transfer is to facilitate knowledge transfer between models with different models, tasks and modalities. Suppose there are two models: a larger capacity model $\mathcal{M}_l$ and a smaller capacity model $\mathcal{M}_s$, each with their own independent datasets $\mathcal{D}_l=\{(x_l^{(i)},y_l^{(i)})\}_{i=1}^{\left | \mathcal{D}_l \right | }$ and $\mathcal{D}_s=\{(x_s^{(i)},y_s^{(i)})\}_{i=1}^{\left | \mathcal{D}_s \right | }$. We denote the weights of two models as $W_l$ and $W_s$, and divide the parameters of each model into two parts: those that participate in knowledge transfer $W_t$ and those that are uninvolved in knowledge transfer $W_u$, i.e., $W_l=\{W_{l,t},W_{l,u}\}$, $W_s=\{W_{s,t},W_{s,u}\}$. We aim to learn a model $\mathcal{M}(\cdot ) $ that can receive these model parameters and generate parameters that integrate knowledge from heterogeneous models, which can be represented as:
\begin{equation}
\{\tilde{W}_{l,t},\tilde{W}_{s,t}\}=\mathcal{M}(\{W_{l,t},W_{s,t}\}),
\end{equation}%
where $\tilde{W}_{l,t}$ and $\tilde{W}_{s,t}$ represent the parameters of the two models after knowledge transfer.

Since the knowledge transfer process from a smaller model to a larger model is structurally symmetrical to the process from a larger model to a smaller model, we only demonstrate the knowledge transfer from the larger model to the smaller model here. 

\subsection{MergeNet}
\label{sec:Sec3.2}
 Initially, we consider using a simple model structure, a Multi-Layer Perceptron (MLP), to transform the parameter matrix of the source model into that of the target model. Specifically, we define the parameter matrices of the two models as $W_{l,t}\in \mathbb{R} ^{N\times M}$ and $W_{s,t}\in \mathbb{R} ^{n\times m}$, respectively. The generation process of $\tilde{W}_{s,t}$ is as follows:
\begin{equation}
\tilde{W}_{s,t}=\xi_2((\xi_1(W_{l,t}))^T),
\end{equation}%
where $\xi_1(\cdot)$ and $\xi_2(\cdot)$ both represent the structures of MLP. However, we notice some issues with using such a simple network structure for knowledge transfer between heterogeneous models: (i) This method directly uses the generated parameters to overwrite the existing parameters of the target model, thereby overlooking the knowledge accumulated in the original parameters of the target model. (ii) The vectors of the generated $\tilde{W}_{s,t}$ are produced independently, which may lead to a loss of information between the vectors.

To address these challenges, we propose MergeNet, which can generate hybrid parameters containing knowledge from heterogeneous models based on parameters from these models, thus elegantly facilitating knowledge transfer between different models. As shown in Figure \ref{fig:model}(a), MergeNet consists of multiple knowledge transfer layers (KTL). Each layer in this architecture receives parameters from the previous layer and produces new parameters based on them:
\begin{equation}
\begin{split}
\{\tilde{W}_{l,t}^{(i)},\tilde{W}_{s,t}^{(i)}\} &= \mathcal{M}_i(\{\tilde{W}_{l,t}^{(i-1)},\tilde{W}_{s,t}^{(i-1)}\}), \\
&\quad\quad\quad\quad\quad\quad\forall i=1,\ldots,L,
\end{split}
\end{equation}
where $\mathcal{M}_i$ represents the $i$-th KTL, $\tilde{W}_{l,t}^{(i)}$ and $\tilde{W}_{s,t}^{(i)}$ respectively denote the parameters generated by the $i$-th KTL, and $L$ signifies the number of layers in the KTL.

\paragraph{Parameter Re-Encode.}Similar to LoRA \cite{hu2021lora}, we re-encode the parameter matrices $W_{l,t}$ and $W_{s,t}$ involved in knowledge transfer through low-rank decomposition:
\begin{equation}
W_{l,t}=B_{l,t}A_{l,t},
\end{equation}%
\begin{equation}
W_{s,t}=B_{s,t}A_{s,t},
\end{equation}%
where $B_{l,t}\in \mathbb{R}^{N\times r}$, $A_{l,t}\in \mathbb{R}^{r\times M} $, $B_{s,t}\in \mathbb{R}^{n\times r}$ and $A_{s,t}\in \mathbb{R}^{r\times m} $ with $r\ll \{N,M,n,m\}$. Our approach significantly differs from LoRA in terms of usage and objectives: (i) LoRA aims to model the incremental updates $\Delta$ of parameters, whereas our method directly applies low-rank decomposition to the parameter matrices themselves. (ii) LoRA decomposes $\Delta$ into two low-rank matrices to reduce the computational cost of fine-tuning. In contrast, we employ low-rank decomposition to obtain manageable low-rank matrices that package the comprehensive knowledge of the original parameters. In our approach, the low-rank matrices are considered as fundamental units containing model knowledge, differing from the potential information loss due to the independence of vectors in MLP-based parameter adapter.

\paragraph{Low-rank Parametric Knowledge Adapter.}MLP-based parameter adapter directly maps the knowledge from the source model to the target model, often overlooking the existing knowledge within the target model. To address this limitation, we introduce Low-rank Parametric Knowledge Adapter (LPKA). This mechanism is used to extract knowledge from the low-rank matrices and merge knowledge from different models to generate new parameters. Figure \ref{fig:model}(c) illustrates the specific process of knowledge transfer through LPKA. Taking into account the knowledge from both models, we flatten $A_{l,t}$ and $A_{s,t}$, obtained from low-rank decomposition, by rows/columns and then apply the attention mechanism to integrate the knowledge from the source model into the target model:
\begin{equation}
\begin{split}
\tilde{A}_{s,t} &= \sum_{i=1}^{4}\omega _i\text{Attn}(\phi_{R/L}(A_{s,t}), \phi_{R/L}(A_{l,t}),\phi_{R/L}(A_{l,t})).
\end{split}
\end{equation}
where Attn represents the softmax attention mechanism, while $\phi_R(\cdot )$ and $\phi_L(\cdot )$ denote the operations of flattening the low-rank matrices by rows and columns, respectively. For the low-rank matrices of both models, there is a choice of flattening  either by rows or by columns, resulting in four possible combinations. Additionally, $\omega_i$ is a learnable parameter used to balance the contribution of the $i$-th part of the attention mechanism.

\paragraph{Training Process.}
During the training process, the optimization updates of a single model through the gradient descent algorithm constitute self-learning, while knowledge transfer between two models involves mutual learning. We believe that only relying on knowledge transfer will not yield optimal results; self-learning should be interleaved with knowledge transfer, akin to a self-consolidation phase under the guidance of a teacher. Therefore, we divide the entire model training process into two phases: the knowledge transfer phase and the self-learning phase. Specifically, we define the knowledge transfer cycle as $T_{cycle}$, during the time step $t$:
\begin{equation}
\tilde{A}_{s,t}=\begin{cases}\mathcal{M}(\{A_{s,t},A_{l,t}\}) 
& \text{if } t \bmod T_{cycle}=0,\\A_{s,t} & \text{otherwise}.\end{cases}
\end{equation}%

We define the loss function of the smaller model as $\mathcal{L}_s$. During the knowledge transfer phase, the parameters $W_s$ of the smaller model are optimized to minimize $\mathcal{L}_s$ with gradient updates: $W_s=W_s-\eta_s\nabla_{W_s}\mathcal{L}_s(\{\tilde{W}_{s,t},W_{s,u}\}) $, where $\eta_s$ is the learning rate. In our work, the parameter adapter is trained concurrently with the network involved in knowledge transfer, and its parameters are defined as $\varphi$. Following the approach of \citet{shamsian2021personalized},  we utilize an update rule:   $\varphi=\varphi -\eta(\nabla_{\varphi}W_{s,t})^T\Delta W_{s,t}$, where $\Delta W_{s,t}$ represents the change after a knowledge transfer cycle. In the testing phase, the parameter adapter is removed, resulting in zero additional overhead.

\section{Experiments}

Our method is designed to facilitate heterogeneous knowledge transfer, independent of the architecture and specific tasks of the models involved. To assess the effectiveness of our approach, we conduct extensive experiments in several challenging scenarios: cross-structure, cross-modal and cross-task.

\subsection{Cross-Structural Knowledge Transfer}
\label{sec:cross_str}
\paragraph{Implementation Details.}
We conduct cross-structure knowledge transfer using the CIFAR-100 \cite{krizhevsky2009learning} dataset. This dataset comprises 100 categories, with training and validation sets containing 50k and 10k images, respectively. We use ResNet50 \cite{he2016deep} as the larger model and MobileNetV2 \cite{sandler2018mobilenetv2} as the smaller model, both of which are not pre-trained.
The goal of cross-structure knowledge transfer is to transfer knowledge from one model's module to a structurally different module in another model. Specifically, we focus on transferring knowledge from the linear classifier of one model to the convolutional layer of another model.

\begin{table}[t]
\setlength{\tabcolsep}{12pt}
    \centering
    \begin{tabular}{ccc}
        \toprule[1.25pt]
        \toprule         Methods
                  & \begin{tabular}[c]{@{}c@{}}Top-1 \\ Acc(\%)\end{tabular} & \begin{tabular}[c]{@{}c@{}}Top-5\\ Acc(\%)\end{tabular} \\
        \midrule
     Vanilla MobileNetV2                       & 63.87  & 88.77 \\
KD \cite{hinton2015distilling}      & 64.32  & 88.62 \\
RKD \cite{park2019relational}    & 65.48  & 88.9  \\
DKD \cite{zhao2022decoupled} &65.23 &89.01 \\
NKD \cite{yang2023knowledge}     & 65.09  & 88.9  \\
MergeNet(R$\rightarrow\!\!\mathrm{M}$)     & 66.23  & 89.66 \\
MergeNet(R$\leftrightarrow\!\!\mathrm{M}$) & \textbf{66.51}  & \textbf{89.75} \\
        \midrule
Vanilla ResNet50                               & 68.11  & 89.61 \\
KD\cite{hinton2015distilling}      & 68.36
  & 89.9 \\
RKD\cite{park2019relational}      & 68.6
  & 90.21  \\
DKD \cite{zhao2022decoupled} &69.03 &90.25 \\
NKD\cite{yang2023knowledge}      & 69.27  & 90.18  \\
MergeNet(R$\leftrightarrow\!\!\mathrm{M}$) & \textbf{69.84}  & \textbf{90.57} \\        
        \bottomrule[1.25pt]
    \end{tabular}
    \caption{Performance of cross-structural knowledge transfer. R and M mean the ResNet50 and MobileNetV2, respectively. R$\rightarrow\!\!\!\mathrm{M}$ denotes knowledge transfer from ResNet50 to MobileNetV2, and R$\leftrightarrow\!\!\mathrm{M}$ represents mutual knowledge transfer between ResNet50 and MobileNetV2. The best results for each setting are highlighted in bold.}
    \label{tab:cifar100}
\end{table}

\paragraph{Cross-Structural Knowledge Transfer Results.} We compare our method against several baselines and present the experimental results on the CIFAR-100 dataset in Table \ref{tab:cifar100}. Our method consistently outperforms the baselines. For instance, on MobileNetV2, our approach achieves a 1.02\% improvement in Top-1 accuracy. Additionally, we investigate the transfer of knowledge from smaller models to larger models. As suggested by \citet{yuan2020revisiting}, student models can enhance the performance of teacher models through reverse distillation. In line with this perspective, we compare traditional knowledge distillation with our method. The results demonstrate that our method surpasses various distillation techniques. The improvement achieved by our method can be attributed to the fact that different models focus on different aspects of a task. Through knowledge transfer, a model can learn the focus points of other models. Moreover, transferring knowledge from the linear classifier to the convolutional layer allows the convolutional layer to learn the focus points of the linear classifier, thereby generating representations more aligned with those of the linear classifier.

\subsection{Cross-Modal Knowledge Transfer}

\paragraph{Implementation Details.}We conduct experiments in cross-modal knowledge transfer on two distinct tasks: Visual Question Answering using the VQA v2.0 \cite{goyal2017making} dataset and Image-Text Retrieval using the MSCOCO \cite{lin2014microsoft} dataset, with X\mbox{-}VLM \cite{zeng2021multi} as the experimental model. For the Image-Text Retrieval task, we use R@10 as the evaluation metric. Given the large size of the datasets and limited computational resources, which led to lengthy training times, we train the model using only 10\% of the training set and assess the effectiveness of cross-modal knowledge transfer on the complete test set. Our objective is to use knowledge from one modality to guide the learning of another modality. Specifically, we explore transferring knowledge from the parameters of the visual encoder to the textual encoder and vice versa.

\paragraph{Cross-Modal Knowledge Transfer Results.}
We conduct cross-modal knowledge transfer experiments in various ways, including unidirectional transfers between modalities and bidirectional transfers where modalities mutually transfer knowledge. The results of these experiments are summarized in Table \ref{tab:Modal}. It is evident from the results that our method provides significant improvements in accuracy across different settings. We speculate that transferring knowledge between modal encoders allows for the integration of different modal information before it enters the modal interactor, thereby easing the process of combining information from various modalities for the interactor.

\begin{table}[t]
\setlength{\tabcolsep}{3.5pt}
\small
\centering
\begin{tabular}{cccc|cc}
\toprule[1.25pt]
\toprule         
    \multirow{2}{*}{Methods}              & \multicolumn{3}{c|}{VQA}        & \multicolumn{2}{c}{ITR} \\
                  & overall   & other     & number & TR         & IR         \\
\midrule
Vanilla        & 45.78     & 31.33     & 28.71  & 41.48      & 37.64      \\
MergeNet(V$\rightarrow\!\!\mathrm{T}$) & 46.33     & 33.29     & 31.33  & 44.72      & 39         \\
MergeNet(T$\rightarrow\!\!\mathrm{V}$) & 45.96     & 31.99     & 31.15  & 44.58      & 38.93      \\
MergeNet & \textbf{46.51}     & \textbf{33.84}     & \textbf{31.54}  & \textbf{44.78}      & \textbf{39.26}      \\      
\bottomrule[1.25pt]  
\end{tabular}
\caption{Performance of cross-modal knowledge transfer. V$\rightarrow\!\!\mathrm{T}$ represents the knowledge transfer from visual to textual modality, while T$\rightarrow\!\!\mathrm{V}$ signifies the transfer from textual to visual modality. The last row indicates the scenario where both visual-to-textual and textual-to-visual knowledge transfers are conducted simultaneously. The best results for each setting are highlighted in bold.}
\label{tab:Modal}
\end{table}

\begin{table}[t]
\setlength{\tabcolsep}{10pt}
\centering
\begin{tabular}{ccc|c}
    \toprule[1.25pt]
    \toprule              \multirow{2}{*}{Methods}
                              & \multicolumn{2}{c|}{SQuAD v2.0} & IMDb      \\
                              & EM$\uparrow$    & F1$\uparrow$ & Err$\downarrow$ \\
    \midrule
Vanilla                    & 70.17         & 73.06         & 8.02      \\
MergeNet                          & 71.89     & 75.43     & 7.5       \\
    \bottomrule[1.25pt]
\end{tabular}
\caption{Performance of cross-task knowledge transfer. BERT executes a question answering task, while DistilBERT performs a sentiment classification task. $\uparrow$($\downarrow$) denotes that a higher (lower) result corresponds to better performance.}
\label{tab:cross_task}
\end{table}

\subsection{Cross-Task Knowledge Transfer}\label{sec:cross_task}

\paragraph{Implementation Details.}
We study the cross-task knowledge transfer effectiveness of our method on the following tasks: a classification task (IMDb sentiment classification \cite{maas2011learning}) and a question answering task (SQuAD v2.0 \cite{rajpurkar2018know}). We utilize BERT \cite{devlin2018bert} and DistilBERT \cite{sanh2019distilbert}, respectively, to perform these tasks. DistilBERT is a distilled version of BERT, maintaining the general architecture of BERT but with half the number of layers. Due to the difference in dataset sizes, we schedule knowledge transfer to occur after every 2$T_{cycle}$ batches in the question answering task and every $T_{cycle}$ batches in the classification task. (unless otherwise specified, knowledge transfer is conducted after every $T_{cycle}$ batches by default).

\paragraph{Cross-Task Knowledge Transfer Results.}
The results of cross-task knowledge transfer on the SQuAD v2.0 and IMDb datasets are shown in Table \ref{tab:cross_task}. Our method achieve performance improvements in both tasks. For example, transferring knowledge from the classification task to the question-answering task results in a 1.72\% increase in Exact Match (EM) score and a 2.37\% increase in its F1 score. Similarly, transferring knowledge from the question-answering task to the classification task leads to a 0.52\% reduction in error rate. We believe that in similar tasks, the forms of knowledge representation are likely similar, and models performing different tasks can enhance their own task performance by learning the knowledge from other tasks. To demonstrate the capability of our method in aligning knowledge across different tasks, we present a more challenging cross-task knowledge transfer scenario in the following section.

\begin{table}
\setlength{\tabcolsep}{2pt}
\centering
\begin{tabular}{ccc|cc}
\toprule[1.25pt]
\toprule
    \multirow{2}{*}{Methods}        & \multicolumn{2}{c|}{CIFAR-100} & \multicolumn{2}{c}{SQuAD v2.0} \\
            & Top-1 Acc     & Top-5 Acc     & EM             & F1            \\
\midrule
Vanilla & 63.87         & 88.77         & 70.17          & 73.06             \\
MergeNet        & 65.56        & 88.74         & 70.89          & 74.15         \\
\bottomrule[1.25pt]  
\end{tabular}
\caption{Performance of challenging cross-task knowledge transfer. MobileNetV2 executes an image classification task and BERT  performs a question answering task.}
\label{tab:Integrated}
\end{table}

\subsection{More Challenging Cross-Task Knowledge}

\paragraph{Implementation Details.}
We conduct cross-structure modal task knowledge transfer experiments in question-answering and image classification tasks. For the question-answering task, we use the BERT model on the SQuAD v2.0 dataset, and for image classification, we utilize MobileNetV2 on the CIFAR-100 dataset. In our approach, we choose to transfer knowledge between the Value matrix of the attention module in the last layer of BERT and the linear classifier of MobileNetV2. As in the previous section, due to the difference in dataset sizes for the two tasks, we adopt a specific strategy to balance the knowledge transfer process. Specifically, we schedule knowledge transfer to occur after every 2$T_{cycle}$ batches in the image classification task and every $T_{cycle}$ batches in the question-answering task.

\paragraph{Integrated Knowledge Transfer Results.}We conduct cross-structure modal task knowledge transfer experiments, which can be seen as a more challenging form of cross-task knowledge transfer. We choose two significantly different tasks for our experiments: question-answering and image classification. As shown in Table \ref{tab:Integrated}, our method is effective in transferring and applying knowledge learned in one task to another, significantly different task. For instance, for MobileNetV2, there is a 1.69\% improvement in Top-1 accuracy, and for BERT, there is a 1.09\% increase in the F1 score. We believe that despite the significant differences between the tasks used in our experiments, there may be some common information processing mechanisms shared among different tasks. By learning knowledge relevant to their own tasks from other tasks, models can improve their performance.

\begin{table}[t]
\centering
\begin{tabular}{ccc}
\toprule[1.25pt]
\toprule
    \multirow{2}{*}{Methods}                    & Top-1          & Top-5          \\
                        & Acc        & Acc        \\
\midrule

Vanilla             & 63.87          & 88.77          \\
Tf-KD \cite{yuan2020revisiting}                   & 65.43          & 88.56          \\
USKD \cite{yang2023knowledge}                    & 65.66          & 86.61          \\
\midrule
Linear Classifier$\rightarrow$IRB-4  & 63.51          & 88.64          \\
Linear Classifier$\rightarrow$IRB-8  & 64.02          & 88.71          \\
Linear Classifier$\rightarrow$IRB-12 & 64.42          & 88.95          \\
Linear Classifier$\rightarrow$IRB-16 & \textbf{66.48} & \textbf{89.49} \\
\bottomrule[1.25pt] 
\end{tabular}
\caption{Performance of self knowledge transfer on the CIFAR-100 dataset. Here, 'IRB-x' denotes the x-th Inverted Residual Block in MobileNetV2. The best results for each setting are highlighted in bold.}
\label{tab:self knowledge transfer}
\end{table}

\begin{table}[t]
    \centering
    \begin{tabular}{ccc}
    \toprule[1.25pt]
    \toprule
        Methods & Params & Top-1 Acc(\%) \\ 
        \midrule
        Vanilla MobileNetV2 & 3.5M & 63.87 \\ 
       Vanilla ResNet50 & 25.5M & 68.11 \\ 
        \midrule
        KD & 29M & 64.32 \\ 
        MergeNet* & 12.1M & 65.92 \\ 
        MergeNet & 37.1M & 66.23 \\ 
    \bottomrule[1.25pt] 
    \end{tabular}
    \caption{Performance of knowledge transfer from a frozen, pre-trained model. * denotes knowledge transfer from a frozen, pre-trained model.}
    \label{tab:pre_train}
\end{table}

\begin{table}[t]
\setlength{\tabcolsep}{12pt}
\centering
\begin{tabular}{ccc}
\toprule[1.25pt]
\toprule  
        Methods           & EM    & F1    \\
\midrule
Layer 6$\rightarrow$Layer 3  & 57.69 & 60.42 \\
Layer 12$\rightarrow$Layer 3 & 54.77 & 58.5  \\
Layer 6$\rightarrow$Layer 6  & 64.89 & 68.26 \\
Layer 12$\rightarrow$Layer 6 & \textbf{66.98} & \textbf{70.44} \\
\bottomrule[1.25pt]
\end{tabular}
\caption{The results of knowledge transfer across different layers on the SQuAD v2.0 dataset. x$\rightarrow\!\!\mathrm{y}$ denotes transferring knowledge from the x-th layer of BERT to the y-th layer of DistilBERT. The best results for each setting are highlighted in bold.}
\label{tab:cross_layer}
\end{table}

\begin{figure}[t]
    \centering
\includegraphics[width=7.5cm]{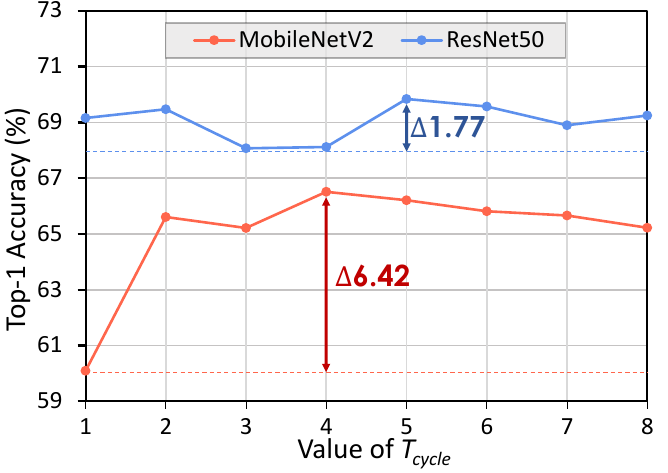}
    \caption{Ablation with respect to $T_{cycle}$.}
    \label{fig:Tcycle}
\end{figure}

\begin{table}[t]
\setlength{\tabcolsep}{2pt}

\centering
\begin{tabular}{ccc}
\toprule[1.25pt]
\toprule
    \multirow{2}{*}{Methods}            &MobileNetV2      &ResNet50       \\
                &Top-1 Acc &Top-1 Acc\\
\midrule
MLP-based       & 64.69 & 68.53 \\
LPKA(Individual Attn) & 65.76 & 69.38 \\
LPKA(Avg Attn)        & 66.02 & 69.66\\
MergeNet        &66.51  & 69.84     \\
\bottomrule[1.25pt]
\end{tabular}
\caption{Ablation study of the effect of individual module.}
\label{tab:ab}
\end{table}

\subsection{Self Knowledge Transfer}

\paragraph{Implementation Details.}To thoroughly evaluate the broad applicability of our method, we conduct a series of self knowledge transfer experiments similar to self-distillation on the CIFAR-100 dataset using MobilenetV2. Specifically, we attempt to transfer knowledge from the linear classifier to the 4th, 8th, 12th, and 16th Inverted Residual Blocks (IRB) out of a total of 17, to test the self-knowledge transfer capability of our method. Furthermore, we compare our approach with state-of-the-art self-distillation methods, including Tf-KD \cite{yuan2020revisiting} and USKD \cite{yang2023knowledge}. These methods obtain additional supervisory signals by setting manual soft labels.

\paragraph{Self Knowledge Transfer Results.}The results of the self-knowledge transfer are shown in Table \ref{tab:self knowledge transfer}. It can be observed that: (i) In terms of single-model self knowledge transfer, our method outperforms existing self-distillation methods, bringing significant improvements to the model. For example, compared to self-distillation methods, the knowledge transfer from the linear classifier to the 16th IRB results in a 0.82\% increase in top-1 accuracy, and a 2.88\% increase in top-5 accuracy. (ii) Transferring knowledge to deeper IRB yields better performance. Deeper blocks have stronger expressive capabilities and can better assimilate the knowledge from the linear classifier. In contrast, shallower blocks struggle to comprehend this knowledge, as evidenced by the less effective transfer to the 4th IRB. (iii) The transfer to the 16th IRB significantly outperforms other settings. This may be because the linear classifier and the 16th block are similar in terms of parameter size and location, leading to comparable mean and variance in their parameters, which facilitates smoother knowledge transfer.

\subsection{Knowledge Transfer from Frozen, Pre-Trained Model}
We attempt to transfer the knowledge from a frozen, pre-trained larger
model to a smaller one. In the setup, the larger model
only needs to retain the parameters involved in the knowledge
transfer and does not require forward propagation, resulting
in fewer computational requirements. Following the settings in the previous section, ResNet50 serves as the pre-trained model. We provide detailed information on the total parameter counts required for each method, denoted as 'Params'. As shown in Table \ref{tab:pre_train}, our approach achieves better performance with lower memory overhead.

\subsection{Knowledge Transfer Across Different Layers}
We explore the effectiveness of our method in transferring knowledge across different layers of information. Similar to the previous section, we use BERT and DistilBERT for our experiments. The results, as shown in Table \ref{tab:cross_layer}, indicate that the most significant performance improvement occurs when both models select their last layers for knowledge transfer. It is widely recognized that deeper layers in neural networks contain more advanced semantic information, enabling a better understanding of higher-level concepts. Therefore, transferring knowledge from these deeper layers imparts richer semantic information to DistilBERT. However, an exception is noted: the transfer between the 6th layer of BERT and the 3rd layer of DistilBERT outperforms the transfer between the 12th layer of BERT and the 3rd layer of DistilBERT. This suggests that while higher-level information typically enhances performance, overly advanced knowledge may be challenging for the model to comprehend, potentially disrupting its existing knowledge structure.

\subsection{Ablation Study}
\paragraph{Knowledge Transfer Cycle  $T_{cycle}$.}
We study the impact of the knowledge transfer cycle $T_{cycle}$, which controls the proportion of self-learning during the training process. Figure \ref{fig:Tcycle} shows the experimental results of knowledge transfer for MobileNetV2 and ResNet50 on the CIFAR-100 dataset under varying $T_{cycle}$ coefficients. We observe that incorporating self-learning during the training process leads to performance improvements. For example, MobileNetV2 achieves a performance of 66.51\% when $T_{cycle}$ is set to 4, representing a 6.42\% increase from the 60.09\% performance without self-learning. 
Additionally, higher self-learning ratios lead to a decline in performance for both models. This is because less frequent knowledge transfer hampers the training of the parameter adapter, reducing the effectiveness of the knowledge transfer. Notably, ResNet50 is less affected by changes in the self-learning ratio compared to MobileNetV2, indicating that larger models are better at absorbing knowledge from smaller models.

\paragraph{Effectiveness of Each Component.}We conduct an ablation study, as shown in Table \ref{tab:ab}, to demonstrate the effectiveness of each component in MergeNet. As mentioned in the previous section, the parameter adapter can alternatively use MLP as the backbone. In this case, the target model directly adopts the knowledge from the source model while ignoring its own accumulated knowledge, potentially leading to training instability. We compare MergeNet with the MLP-based parameter adapter (row 1 vs
row 4). The results show that the MLP-based parameter adapter significantly decreases performance by 1.82\% and 1.31\% for both models, respectively. Furthermore, we compare MergeNet with different variants of LPKA: (i) LPKA(Individual Attn), which uses only the row-flattened form of the low-rank matrix; (ii) LPKA(Avg Attn), which averages each softmax attention module without using trainable weight parameters. The results indicate that using only the row-flattened form is less effective than considering both row and column, suggesting that column vectors in the parameter matrix also contain crucial model knowledge. Additionally, the performance of MergeNet may decline without trainable weight parameters. These results validate the contribution of each component to enhancing model performance.

\section{Conclusion}
We propose a novel knowledge transfer framework named MergeNet. This framework utilizes model parameters as the natural carriers of knowledge for the transfer process, independent of specific model architectures and tasks. MergeNet adaptively extracts the required knowledge from the source model based on the knowledge needs of the target model. We hope that MergeNet will provide new insights into the field of knowledge transfer.

\section{Acknowledgements}
This work was supported by the National Natural Science Foundation of China (No. 62441605, 62402429), the National Science and Technology Major Project (No. 2022ZD0119100) and the Key Research and Development Program of Zhejiang Province (No. 2024C03270). This work was also partially supported by the ZJU Kunpeng\&Ascend Center of Excellence and the Ningbo Yongjiang Talent Introduction Programme (No. 2023A-397-G).

\bibliography{aaai25}

\end{document}